\documentclass[11pt,letterpaper]{article}
\usepackage{authblk}
\usepackage[hyperref]{acl2021}
\usepackage{times}
\usepackage{latexsym}

\usepackage{microtype}
\usepackage{subfig}
\usepackage{booktabs}
\usepackage{enumitem}
\usepackage{multirow}
\usepackage{graphicx}
\usepackage{float}
\usepackage{comment}
\usepackage{soul}
\usepackage[ruled,vlined]{algorithm2e}
\usepackage{balance}
\usepackage{adjustbox}
\setlist{nosep}

\aclfinalcopy 
\def\aclpaperid{3298} 

\title{On the Distribution, Sparsity, and Inference-time Quantization of Attention Values in Transformers}

\author[1]{\bf Tianchu Ji}
\author[2]{\bf Shraddhan Jain}
\author[2]{\bf Michael Ferdman}
\author[1]{\\ \bf Peter Milder}
\author[2]{\bf H. Andrew Schwartz}
\author[2]{\bf Niranjan Balasubramanian}
\affil[1,2]{Stony Brook University\protect\\\vspace{-1em}}
\affil[1]{\{tianchu.ji, peter.milder\}@stonybrook.edu}
\affil[2]{\{shrjain, mferdman, has, niranjan\}@cs.stonybrook.edu}

\date{}

\begin{document}
\maketitle
\begin{abstract}

How much information do NLP tasks really need from a transformer's attention mechanism at application-time (inference)?
From recent work, we know that there is sparsity in  transformers and that the floating-points within its computation can be discretized to fewer values with minimal loss to task accuracies. 
However, this requires retraining or even creating entirely new models, both of which can be expensive and carbon-emitting. 
Focused on optimizations that do not require training, we systematically study the full range of typical attention values necessary. 
This informs the design of an inference-time quantization technique using both pruning and log-scaled mapping which produces only a few (e.g. $2^3$) unique values. 
Over the tasks of question answering and sentiment analysis, we find nearly 80\% of attention values can be pruned to zeros with minimal ($< 1.0\%$) relative loss in accuracy. 
We use this pruning technique in conjunction with quantizing the attention values to only a 3-bit format, without retraining, resulting in only a 0.8\% accuracy reduction on question answering with fine-tuned RoBERTa.
\end{abstract}

\section{Introduction}
\label{sec:intro}
While the verdict is still out on which large language model will prove best, at this point in time, all contenders rely on multi-headed attention over multiple layers.
Many have investigated whether attention (the output of the softmax, $\alpha$) itself is \textit{qualitatively} sensible (e.g., correlating with linguistic aspects)~\cite{vig_analyzing_2019,clark_what_2019,voita_context-aware_2018,voita_analyzing_2019,kovaleva_revealing_2019, rogers_primer_2020} or how useful it is for interpreting models~\cite{jain_attention_2019, wiegreffe_attention_2019,brunner_identifiability_2019,rogers_primer_2020}.
Others have focused on inducing sparsity in the attention: 
whether some of the structural components (the softmax function, attention heads and layers) introduce attention sparsity \cite{correia_adaptively_2019,michel_are_2019,voita_analyzing_2019,sajjad_poor_2020}, if the model tends to focus on a small amount of tokens~\cite{clark_what_2019,ramsauer_hopfield_2020}, and the interpretability of such sparsity \cite{chen_lottery_2020,rogers_primer_2020}.
Yet, little is known about our ability to induce sparsity or reduce its values \textit{at application-time}, and what role the inherent sparsity could play in building inference-time efficient transformers.

This work focuses on a systematic study of the quantitative distribution of the attention values across the layers and heads as well as the potential for reducing the information content of attention values during inference at application-time\footnote{Our analyzing code and data are available at \scalebox{0.9}{\url{https://github.com/StonyBrookNLP/spiqa}}}. 
We consider two popular pretrained transformer models: BERT~\cite{devlin_bert_2019} and RoBERTa~\cite{liu_roberta_2019} over tasks of Masked Language Modeling as well as question answering and sentiment analysis.
We explore the attention distributions on the different models and tasks, and quantitatively profile the sparse attention that commonly exists in the transformer model.
Motivated by the high levels of inherent sparsity in these distributions, we design a pruning and quantization technique and test the limits of information necessary from attention. 
 
We find that most attention values can be pruned (i.e. set to zero) and the remaining non-zero values can be mapped to a small number of discrete-levels (i.e. unique values)  without any significant impact on accuracy. Approximately 80\% of the values can be set to zero without significant impact on the accuracy for QA and sentiment analysis tasks.
Further, when we add quantization utilizing a log-scaling, we find a 3-bit discrete representation is sufficient to achieve accuracy within 1\% of using the full floating points of the original model. 

\section{Method}
\label{sec:method}

To analyze attention distribution we first plot histograms of attention values for BERT~\cite{devlin_bert_2019} and RoBERTa~\cite{liu_roberta_2019} models. We also compute a \emph{sparsity distribution} using the proportion of the attention values smaller than a given threshold. For attention pruning, we find attention values that are below a specified threshold and replace them with zero. We experiment with different thresholds. For quantization to $k$-bits we map the continuous attention values to one of $2^k$ real values\footnote{Note here we use full precision floating point rather than a $k$-bit value since our main goal is to see how many discrete levels of attention is needed.}. We use two methods: (i) Linear - Bin the attention values to $2^k$ quantiles and set the mid-point of each as the quantized value. (ii) Log - Bin the log transformed attention values and pick the mid-point of each on the log scale as the quantized value. The quantization methods are explained in detail in Appendix~\ref{appx:quant_method}.

We apply these inference-time (i.e. no training) techniques on three tasks: masked language modeling, question answering and sentiment analysis. For QA we used BERT\footnote{{\scalebox{0.66}{\url{http://huggingface.co/csarron/bert-base-uncased-squad-v1}}}} and RoBERTa\footnote{{\scalebox{0.66}{\url{http://huggingface.co/csarron/roberta-base-squad-v1}}}} models fine-tuned on SQuAD v1.1~\cite{rajpurkar_squad_2016}. For sentiment analysis we used RoBERTa\footnote{\scalebox{0.66}{\url{http://huggingface.co/textattack/roberta-base-SST-2}}} fine-tuned on the SST-2 dataset~\cite{socher_recursive_2013}. For both these tasks we report accuracy on the corresponding development sets. For the Masked Language Modeling (MLM) task we report pseudo-perplexity \cite{salazar_masked_2020} computed on the Huggingface Wikipedia dataset\footnote{\scalebox{0.66}{\url{https://huggingface.co/datasets/wikipedia}}}.

\section{Evaluation}
\label{sec:obs}

\paragraph*{Attention distribution and sparsity.}
\label{subsec:overview}
\begin{figure*}[!ht]
    \centering
    \subfloat[][Layer 1 Head 4 \label{fig:overview_a}]
    {\includegraphics[width=0.33\linewidth]{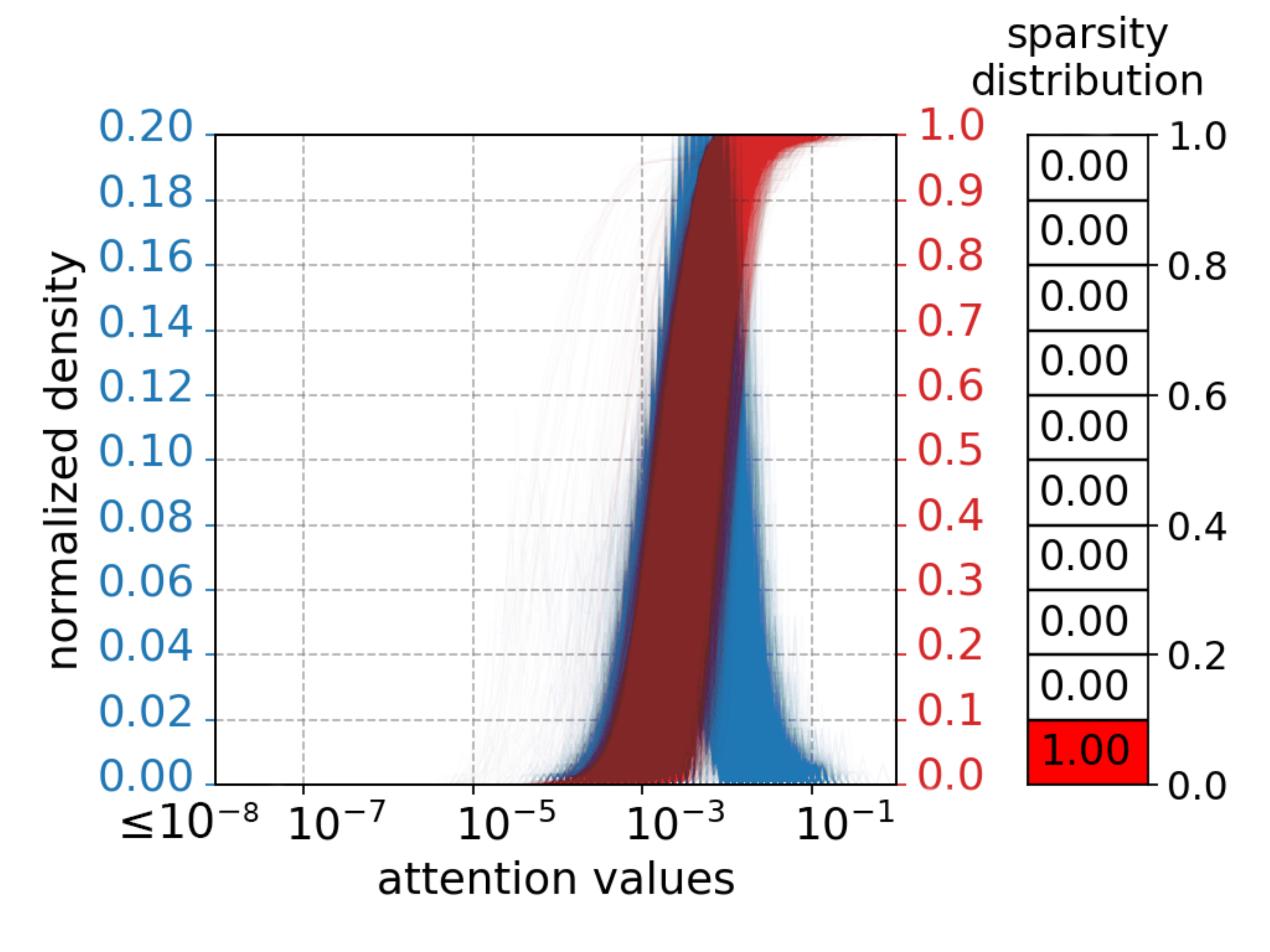}}
    \subfloat[][Layer 2 Head 3 \label{fig:overview_b}]
    {\includegraphics[width=0.33\linewidth]{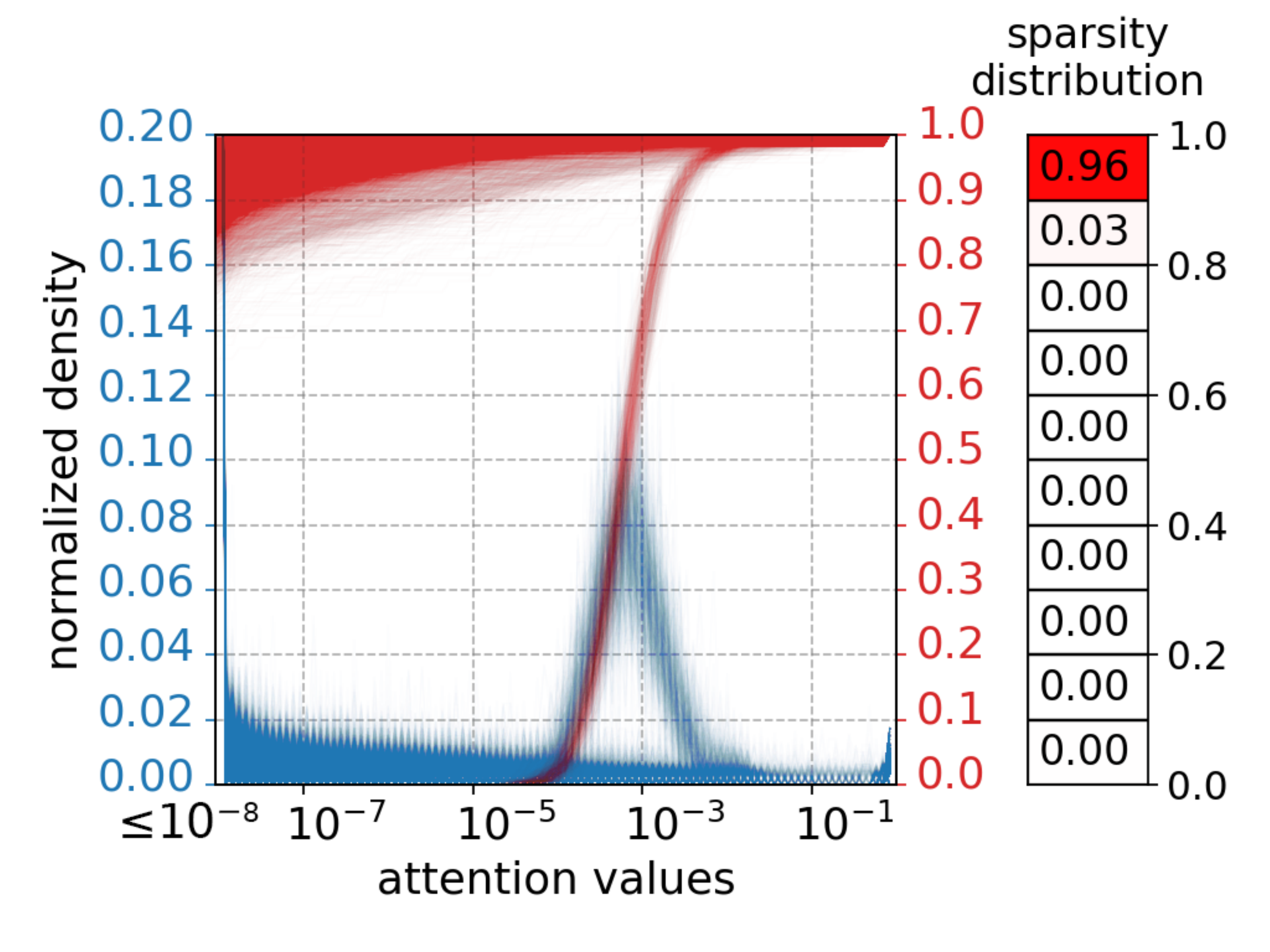}} 
    \subfloat[][Layer 12 Head 11 \label{fig:overview_d}]
    {\includegraphics[width=0.33\linewidth]{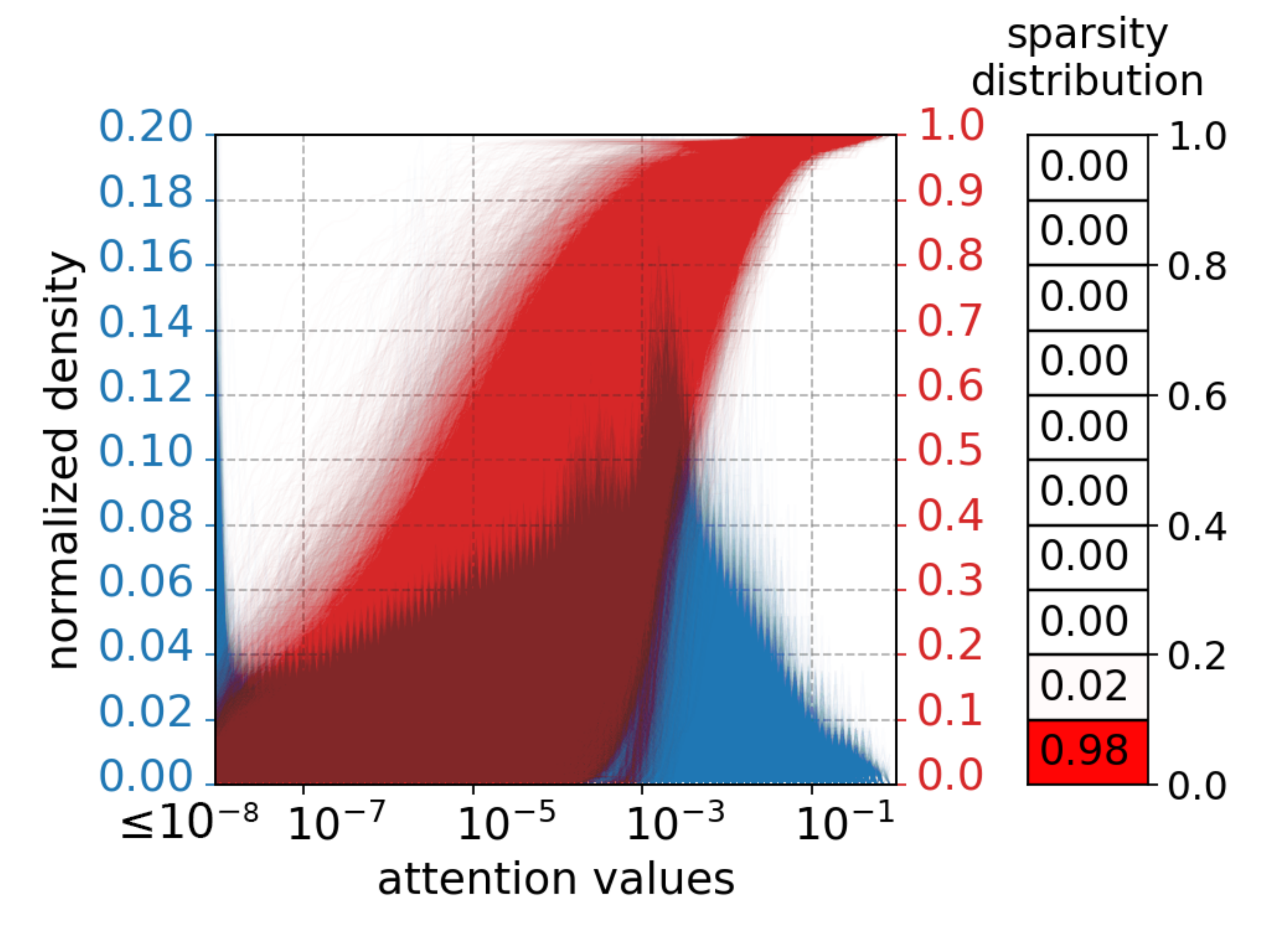}}
    \caption{Normalized histograms (in blue) and cumulative histograms (in red) for every token's attention to others ($\alpha_i$) at different heads in the pretrained RoBERTa model, starting from $10^{-8}$.
    The histograms show different patterns of attention distribution.
    E.g., in (b) many tokens’ attention form an evenly distributed histogram from $10^{-8}$ to 1, and most of the $\alpha_i$ have 80\%--100\% of all the attention values ($\alpha_{ij}$)  $\leq 10^{-8}$. This indicates a higher level of sparsity compared to (a) and (c). 
    The ``sparsity distribution'' bar on the right shows the density of $\alpha_i$ to each level of sparsity. 
    E.g., the red cell with ``0.96'' between 0.9--1.0 in (b) means 96\% of all $\alpha_{i}$ have sparsity between 90\%--100\%, whereas the sparsity is the proportion of $\alpha_{ij}$ in $\alpha_i$ that are less than $10^{-8}$.}
    \label{fig:overview}
\end{figure*}

A thorough quantitative analysis on the attention distribution could help build efficient transformers by providing useful information, such as the degree of sparsity and the range of the attention values.
We plot the histogram of each token’s attention to all the others ($\alpha_i$) and provide three examples of the heads in Figure~\ref{fig:overview} to investigate the density of the attention values, how differently the tokens attend to others in the same attention head, and how sparse a token/head/layer's attention can be.
We find that, for most of the heads, attention forms a lognormal-like distribution similar to Figure~\ref{fig:overview_a}.
On some heads, some of the attention for query token ($\alpha_i$) have more tiny attention values ($\alpha_{ij}$) and induce more sparsity than others (like in Figure~\ref{fig:overview_d}). 
We also observe entire heads with high sparsity, in which nearly all tokens only slightly attend to others (like in Figure~\ref{fig:overview_b}).
Our observation confirms the existence of sparsity in the attention heads.

A key motivation for us is to quantitatively characterize sparsity, especially in terms of how much potential there is in reducing the information content in attention values. 
To this end, we specifically measure the proportion of small attention values by counting the number of $\alpha_{ij}$ that sum up to 0.5 in each $\alpha_i$. 
This indicates that most heads focus strongly on fewer than 10 tokens on average (details in Appendix~\ref{appx:bert_consis}), leading to notable sparsity and suggesting large potential for conveying the same information as continuous attention values using fewer discrete levels.

Beyond these, we occasionally observe outlier attention histograms (like the outliers between $[10^{-4}, 10^{-1}]$ in Figure~\ref{fig:overview_b}).
We also found noticeable differences on the attention histograms from layer to layer.
These findings are related to the works on the syntactic heads/special tokens~\cite{voita_analyzing_2019, kovaleva_revealing_2019, voita_context-aware_2018, clark_what_2019, rogers_primer_2020}) and the differences of the layers/heads~\cite{correia_adaptively_2019, clark_what_2019}.
We discuss how our findings relate to them in Appendices~\ref{appx:bert_diffusion} and \ref{appx:outlier}.

\paragraph*{Limited effect of near-zero attention values during inference.}
\label{subsec:acc_vs_spars}
\begin{figure*}[!ht]
    \centering
    \includegraphics[width=0.98\textwidth]{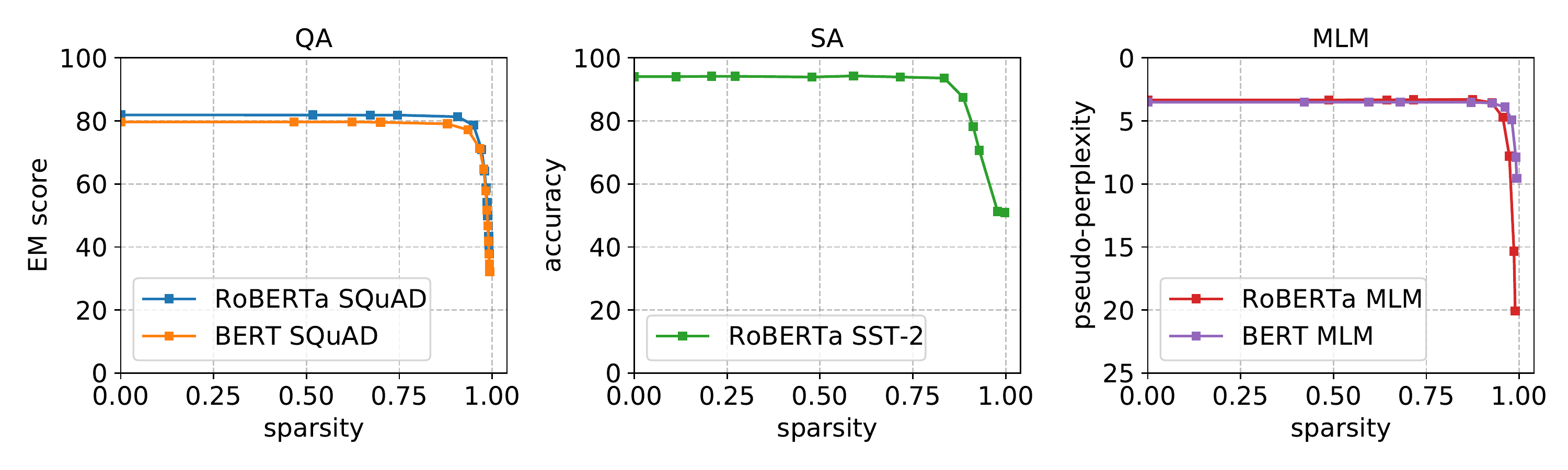} \label{fig:perf_spars_model}
    \caption{Exact Match score (for QA), Accuracy (for SA) and pseudo-perplexity (for MLM) under different levels of sparsity that we induce, showing that on these models and tasks $\sim$80\% of the sparsity can be induced with limited performance drop. X-axis values denotes the induced sparsity levels measured as the proportion of the attention values less than a specified threshold.}
    \label{fig:perf_vs_sparsity}
\end{figure*}

The inherent sparsity we observed motivates us to explore the sparsity of attention at inference-time---how much attention can be pruned during inference, without impacting the model accuracy?
By setting up a series of pruning thresholds, we clamp different proportions of the attention to zero and examine how attention sparsity affects the accuracy, on both pretrained and fine-tuned models.
The results shown in Figure~\ref{fig:perf_vs_sparsity} indicate that the sparsity can grow above 80\% with only a 0.1\%--1.3\% drop in accuracy. 
Specifically, the pretrained BERT model achieves 99.9\% of the original performance with 87\% of the sparsity on Masked Language Modeling. 
By comparing RoBERTa's accuracy on different tasks, we find that sentiment analysis suffers more from increased sparsity, suggesting that different models are differentially sensitive to the induced sparsity.
Our results quantitatively show how much sparsity can be induced in all the attention values without losing accuracy, suggesting that one can expect to prune up to 80\% of the attention values without retraining.

\paragraph*{Quantizing pruned attention.} 
\label{subsec:quantize}

\begin{figure*}[!ht]
    \centering
    \subfloat[][EM scores of the models with differently quantized attention \label{fig:perf_vs_quant}]
    {\includegraphics[height=9.8em]{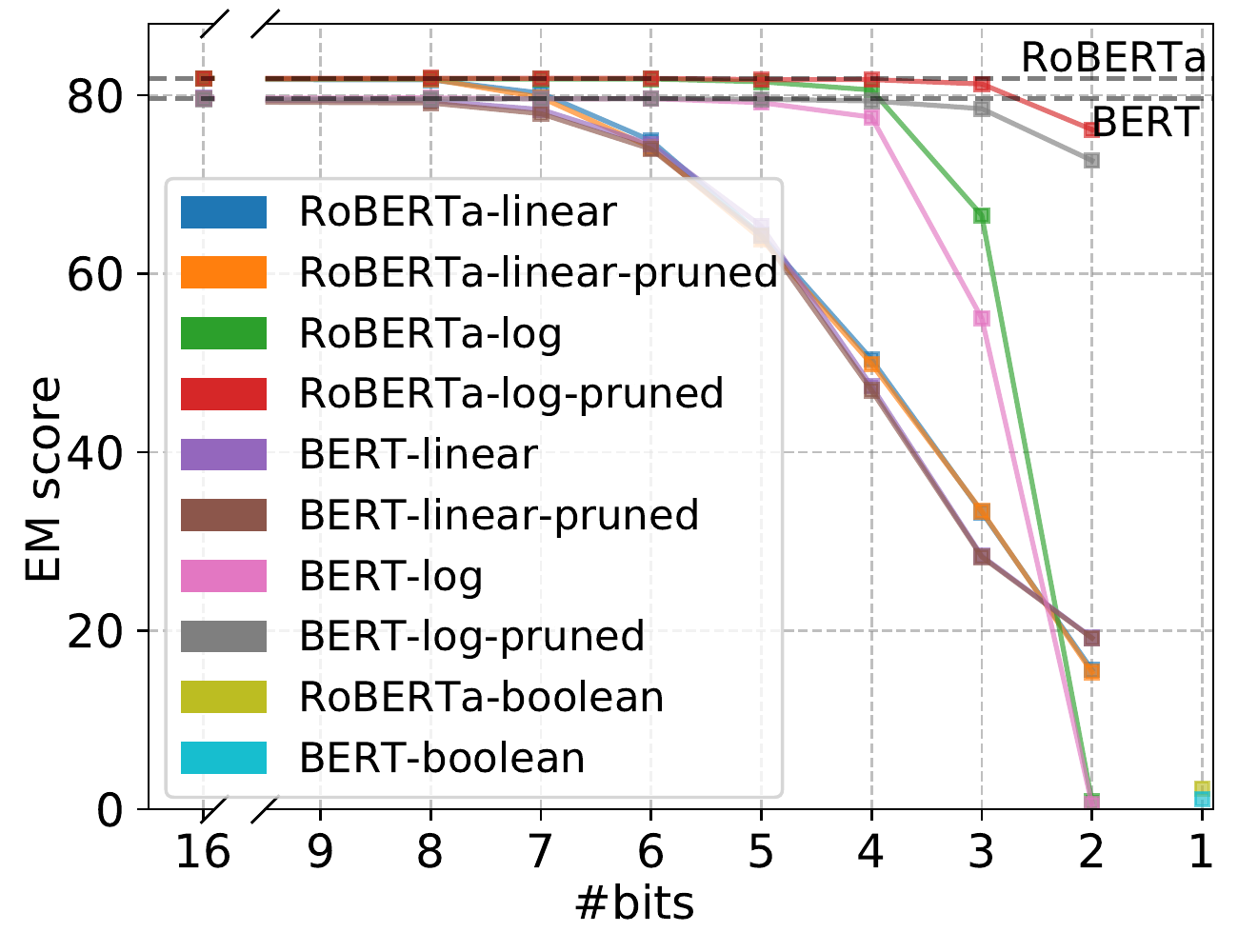}}
    \hspace{1em}
    \subfloat[][performance with different pruning thresholds for 2-bit log quantization \label{fig:clamp_sweep}]
    {\includegraphics[height=9.8em]{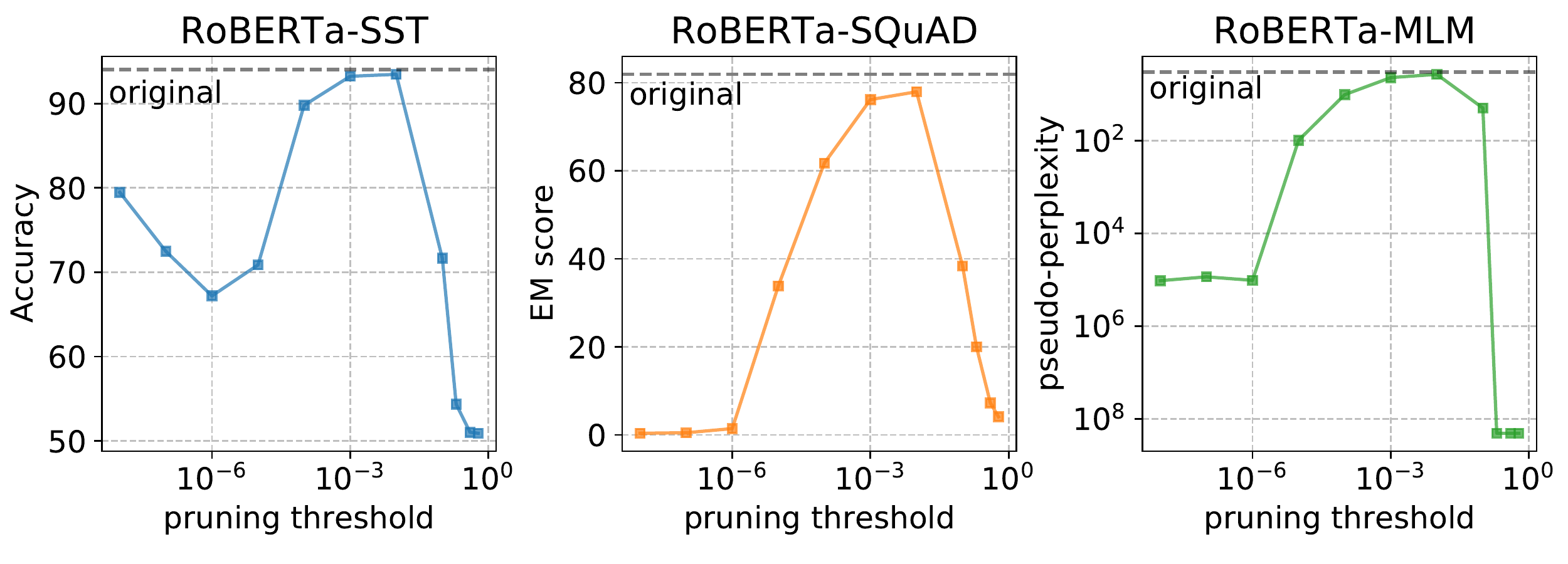}}
    \caption{Performance of the quantized models with/without attention pruning, showing that the attention can be effectively quantized to as low as 3 bits with certain pruning thresholds.
            (a) Exact Match scores for the QA with different quantization methods on fine-tuned BERT and RoBERTa. ``Boolean'' quantization is provided as the extreme case of quantization to a single bit. 
            The pruning has only negligible effect on the linear scale quantization so that ``*-linear'' and ``*-linear-pruned'' curves are highly overlapped.
            (b) Accuracy of the fine-tuned RoBERTa models with 2-bit quantized attention for QA, SA and MLM respectively.
            The attention is pruned before quantization by using different thresholds (shown on the x-axis).
            In all the figures, the original model's performance scores are marked with black dashed lines.}
    \label{fig:quant}
\end{figure*}

Quantization is often used to compress transformer models for higher computational and memory efficiency. Recently \citet{prato_fully_2020} showed that for machine translation, attention values in transformers can be quantized with only a small impact on accuracy. While their results suggest that full precision attention values may not be necessary for high accuracy, it is unclear if one can retain the accuracy in inference-time quantization in general settings i.e., without retraining.
\citet{bhandare_efficient_2019, shen_q-bert_2020, prato_fully_2020} have proved the importance of meticulously selecting the range of the quantization when pursuing higher accuracy.
Intuitively, pruning the tiny attention values will lead to a narrower quantization range with more precise value representatives.
For example, if all $\alpha < 10^{-3}$ are pruned before 3-bit quantization, all numbers we need to quantize will land in $[10^{-3},1]$ rather than $[0,1]$, with the 8 quantiles of the quantization located more densely; this forms a higher resolution within the quantization range compared to the non-pruned version.
Since we observed that pruning most of the attention values during inference has minimal effect on the accuracy when removing only the tiny attention values ($\alpha < 10^{-3}$ in our case),
we hypothesize that properly pruning attention values will help increase the accuracy of the quantized model.

To verify the pruning hypothesis, we selected two quantization methods: linear scale quantization and logarithmic scale quantization (details in Appendix~\ref{appx:quant_method}), quantized only the transformers' attention with various number of bits, and measured the accuracy of the models.
Then we repeated the experiment but pruning $\alpha < 10^{-3}$ (which creates $\sim$80\% sparsity with limited accuracy drop in our sparsity experiment) before quantizing the attention. 

We evaluate the models on different tasks to compare how pruning the attention affects the accuracy when quantizing.
Results in Figure~\ref{fig:perf_vs_quant} show that for both BERT and RoBERTa models, log quantization is greatly improved after pruning, especially with the 3-bit and 2-bit quantization.
Notably, the 3-bit log quantization with pruning only loses 0.8\% and 1.5\% of the original accuracy for the RoBERTa and BERT, respectively.
Contrarily, the pruning has very limited effect on the linear quantization because the selected pruning threshold results only in a negligible change to the effective quantization range. (Details are provided in Appendix \ref{appx:linear_quant_prune}.)
We also repeated the experiment on other tasks and found 2-bit log quantization with pruning only loses 0.7\% accuracy on RoBERTa fine-tuned for sentiment analysis. (Full results are provided in Appendix~\ref{appx:quant_sa_mlm}.)

We further experimented with different pruning thresholds (Figure~\ref{fig:clamp_sweep}) and observed that pruning $\alpha<10^{-2}$ gives the best performance; the threshold can undermine model accuracy if it is either too large ($>10^{-2}$) or too small ($<10^{-3}$).

Our results prove that pruning the sparse attention values helps recover model accuracy with log-scale quantization methods, without any retraining or fine-tuning. 
With attention pruning, a transformer can retain a comparable amount of accuracy even with a simple, low-precision quantized attention (in our case, a 3-bit log quantization). 

\paragraph*{Discussion.} 
\label{sec:benefits}

Sparsifying the attention can help reduce both the computation and memory cost of self-attention during inference. 
Our experiments above demonstrate that it is possible to prune approximately 80\% of attention values while quantizing them to a 3-bit representation. 
Specialized hardware (FPGA and ASIC) can be designed to efficiently operate on highly quantized datatypes and to ``skip'' the zeros to accelerate deep learning inference, such as \citet{albericio_cnvlutin_2016} (which targets CNNs).
Our results show that such an accelerator could effectively reduce the arithmetic cost of computing attention matrices by 80\% and reduce the memory footprint of the attention matrices by up to 96\% (compounding the effect of sparse representation and quantization).
Although attention matrices are not occupying a huge amount of storage, these memory savings can potentially greatly increase the efficiency of a specialized hardware accelerator by reducing its on-chip SRAM usage and/or its memory bandwidth requirement.
Further, the computational savings can help reduce the latency.
Lastly, it is important to note that the benefits of attention sparsity may extend much further than just computing attention values themselves; other computations in the transformer network can also benefit from leveraging the high degree of sparsity without retraining/fine-tuning, potentially yielding larger benefits.
Future work will investigate the computational benefits of utilizing attention sparsity and the design of customized hardware accelerators to efficiently do so.

\section{Related Work}
\label{sec:related}
\paragraph*{Attention distribution.} 
Many have abstractly studied the attention distribution from different aspects~\cite{clark_what_2019, pascual_telling_2021, ramsauer_hopfield_2020, correia_adaptively_2019}, but none specifically have shown the histogram of the $\alpha_i$ directly, 
nor did they investigate the sparse attention values quantitatively. 
\citet{correia_adaptively_2019} indicated that not all of the sparsity in attention was caused by the softmax, and it remained unclear whether such sparsity affected accuracy (which is inspected in this paper).

\paragraph*{Pruning.}
\citet{voita_analyzing_2019, sajjad_poor_2020, michel_are_2019, kovaleva_revealing_2019} pruned one or more heads/layers resulting in comparable or higher model accuracy, either with or without fine-tuning. 
These approaches assume that some heads/layers interpret the information redundantly, which is not always true~\cite{brunner_identifiability_2019, rogers_primer_2020}.
In contrast, our work focuses on a more general method of inducing attention sparsity without operating at layer/head granularity.

\paragraph*{Quantization.} 
\citet{bhandare_efficient_2019, shen_q-bert_2020, prato_fully_2020} have shown benefits from selecting the quantization range, which motivates us to prune the attention before quantization (Section~\ref{subsec:quantize}).
\citet{kim_i-bert_2021, zafrir_q8bert_2019, prato_fully_2020} required re-training while ours does not.
\citet{zhang_ternarybert_2020, bai_binarybert_2020, zadeh_gobo_2020} focused on quantizing the weights rather than the attention values, which is out of our scope.

\paragraph*{Sparse transformers and attention visualization}  
\citet{parmar_image_2018, child_generating_2019, ho_axial_2019, beltagy_longformer_2020, ravula_etc_2020, li_big_2019,tay_efficient_2020} have proposed/summarized various kinds of efficient transformers utilizing induced attention sparsity. 
However, none of them quantitatively analyzed the statistical distribution and the tiny values of the attention.
\citet{vig_multiscale_2019, hoover_exbert_2020} proposed instance-level attention visualization tools.
These are complementary to our quantitative visualization of the distributions of all attention values.

\section{Conclusion}
\label{sec:discussion}
We demonstrated that pruning near-zero values and large reductions in the number of bits needed for attention, even at application time without retraining or fine-tuning, is possible with little loss of accuracy.
This suggests attention plays a very coarse role in model accuracy at inference-time, yielding opportunities to run transformers more efficiently over applications. 
While quantization during training had previously shown promise (down to three bits, for most weights of the transformer), 
we observed the same reduction potential on attention values at application-time, allowing their representation to be reduced down to three bits (or even two for sentiment) with little effort (e.g., without retraining or using a dynamic quantization range).
This shows it is feasible to implement efficient transformers by leveraging heavily sparse and quantized attention values, suggesting the possibility of building specialized hardware (e.g., FPGA and ASIC accelerators) to optimize the transformer’s evaluation on-the-fly.

\section*{Acknowledgments}
We would like to express our appreciation to Adithya V. Ganesan who assisted with our experiments.

This material is based upon work supported by the National Science Foundation under Grant Nos. 2007362 and 1918225.
The experiments were conducted with equipment purchased through NSF Grant No. OAC-1919752.

\bibliography{acl21_related_works}
\bibliographystyle{acl_natbib}

\clearpage
\appendix
\section{Consistency of Inducing Sparsity in the Attention}
\label{appx:bert_consis}

\begin{figure*}[!ht]
    \centering
    \subfloat[][RoBERTa]
    {\includegraphics[width=\linewidth]{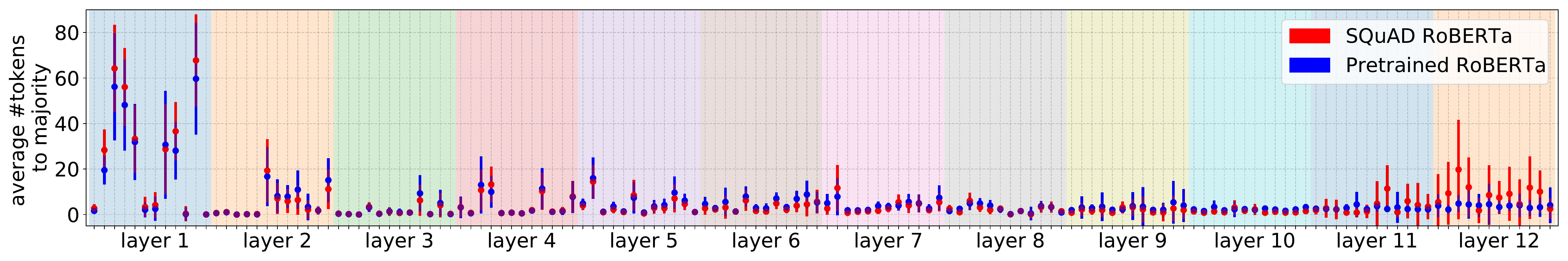} \label{fig:head_consistency_roberta}} \\
    \subfloat[][BERT]
    {\includegraphics[width=\linewidth]{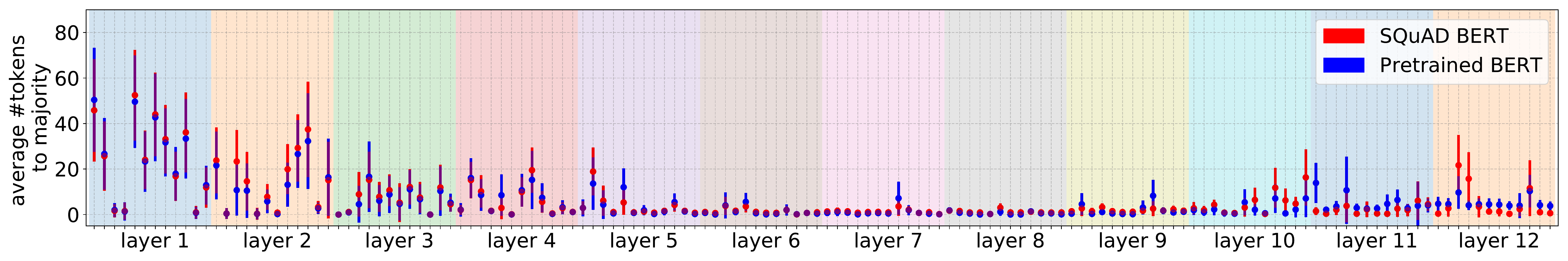} 
    \label{fig:head_consistency_bert}}
    \caption{Mean and standard deviation of the number of tokens' attentions needed to cover a majority (i.e. sum to 0.5) of attention densities in both pretrained and SQuAD-fine-tuned RoBERTa/BERT models.
    Different layers are distinguished by different colors.
    In each layer the error bar represents the mean and std of head 1, head 2, ... , head 12 from the left to the right respectively.}
    \label{fig:head_consistency}
\end{figure*}

Because the softmax function normalizes its input into a probability distribution that sums to 1 and larger values are projected to larger probabilities,
when highly focused tokens with close-to-one probability appear in the attention, they must be accompanied by a large number of near-zero attention values like in Figure \ref{fig:overview_b}.
Thus, the number of close-to-one attention values not only represents how many tokens are strongly attended, but also whether $\alpha_i$ has many near-zero attention values.

To quantitatively evaluate the proportion of these tiny attention values, we computed the number of the largest values in each $\alpha_i$ that sum to 0.5, visualizing their mean and standard deviation in Figure~\ref{fig:head_consistency}. 
On both pretrained RoBERTa and SQuAD-fine-tuned RoBERTa, we observed that most of the heads require on average fewer than ten attention values to sum up to 0.5, meaning that most heads focus strongly on fewer than ten tokens on average, leading to notable sparsity.
We observe that seven of twelve heads in the first layers of both models have a larger average number ($> 10$) of such major tokens. 
For deeper layers, the average number of major tokens decreases. 
Finally, in the last two layers, we again see an increasing trend in the average number of major tokens.
This indicates that middle layers commonly focus on only a small number of tokens, making these layers rich in sparsity.
This confirms the ``sparse deeper layers'' identified by \citet{correia_adaptively_2019, clark_what_2019} and further proves the existence of heavily focused tokens.
It implies the large potential of inducing sparsity in the transformers and motivates us to explore how these sparse attention values contribute to the model accuracy.
We also examined the BERT pretrained model and SQuAD-fine-tuned model, and we found behavior similar to RoBERTa.
Figure~\ref{fig:head_consistency} shows the average of major tokens in the pretrained BERT and SQuAD-fine-tuned BERT.

\section{Dispersion of Attention Histograms}
\label{appx:bert_diffusion}
\begin{figure*}[!ht]
    \centering
    \subfloat[][Layer 1 head 1]
    {\includegraphics[width=0.4\linewidth]{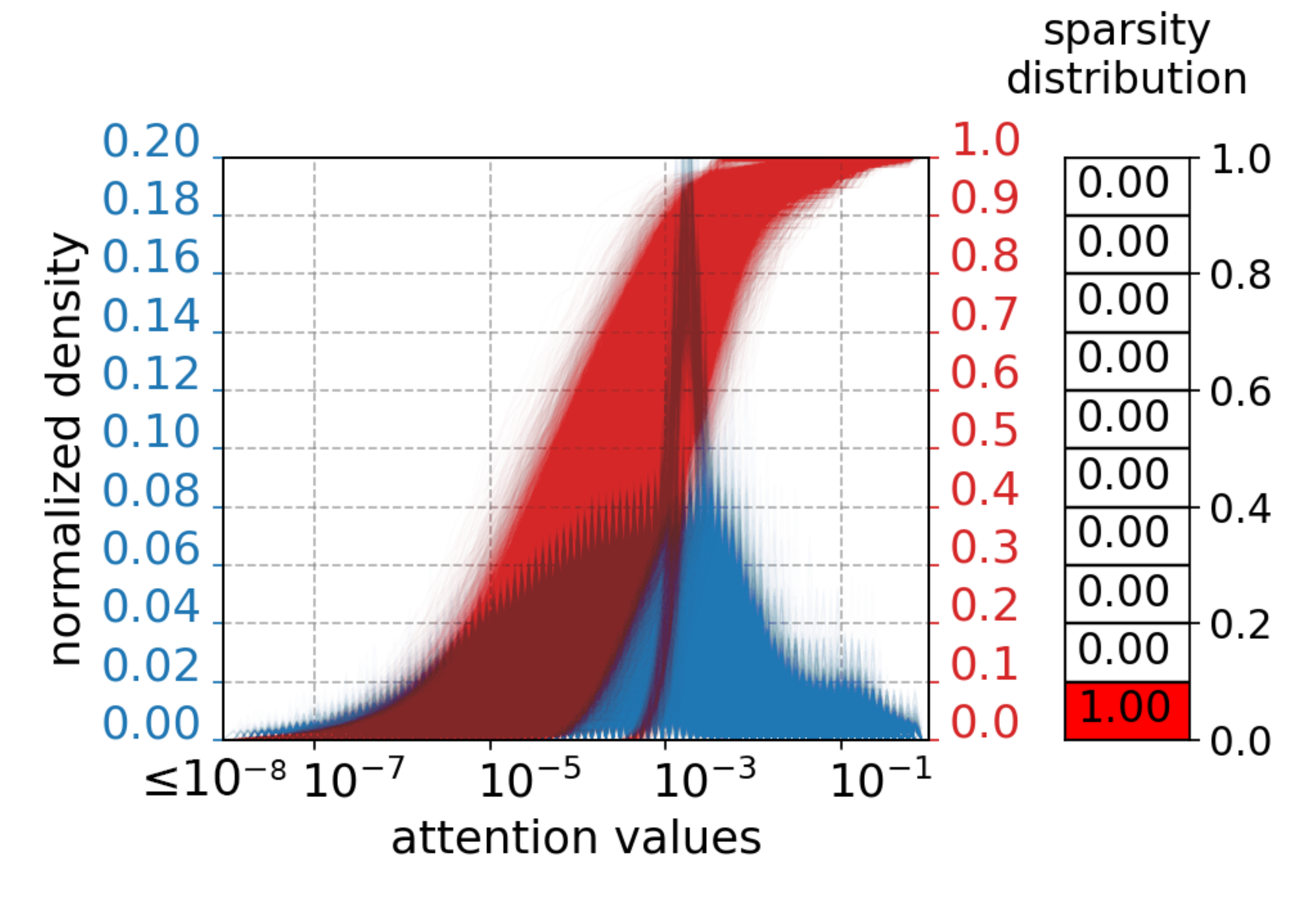} \label{fig:spread_a}}
    \hspace{0.5em}
    \subfloat[][Layer 12 head 1]
    {\includegraphics[width=0.4\linewidth]{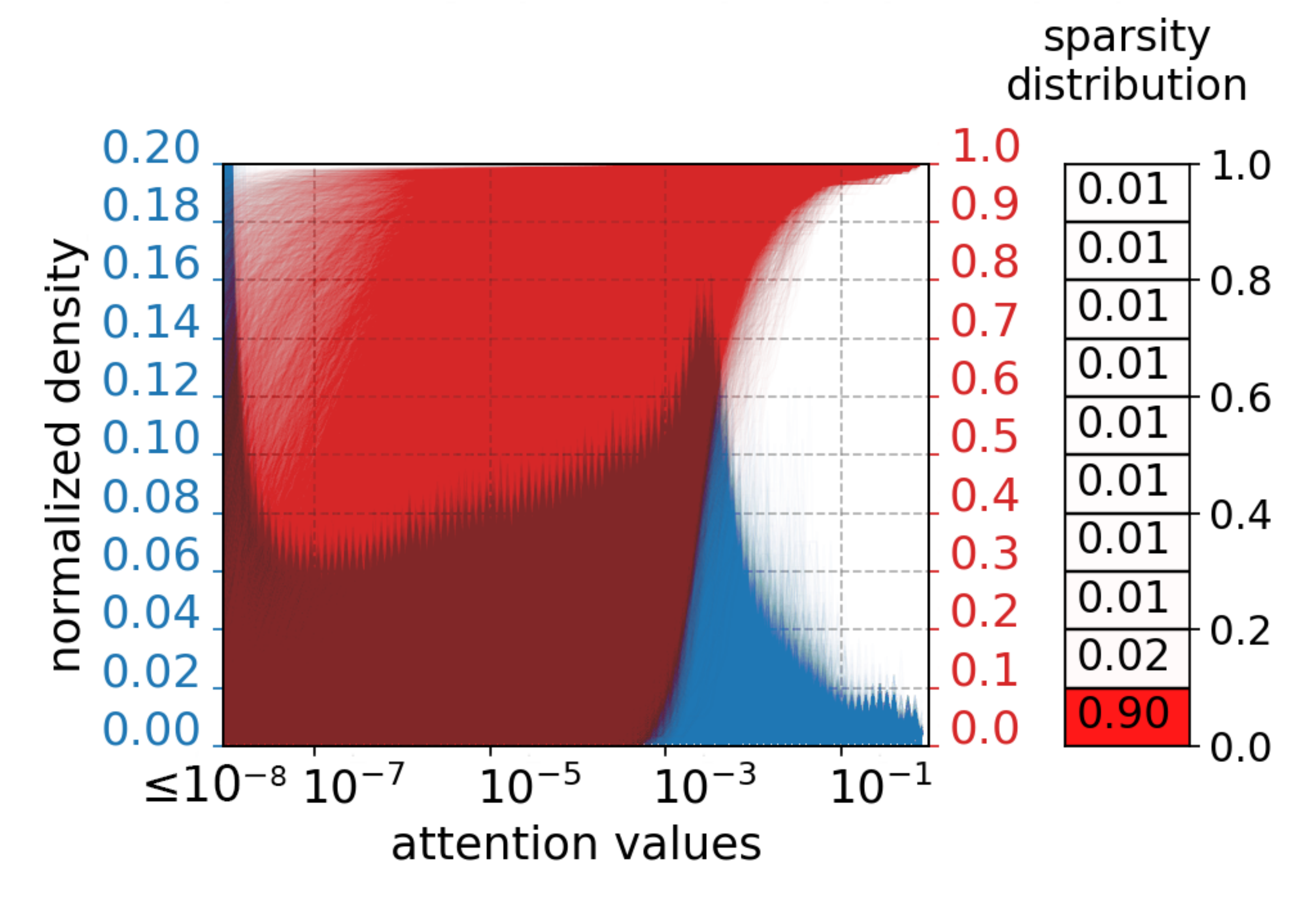} \label{fig:spread_b}} \\
    \subfloat[][Average dispersion of attention per layer in RoBERTa]
    {\includegraphics[width=0.4\linewidth]{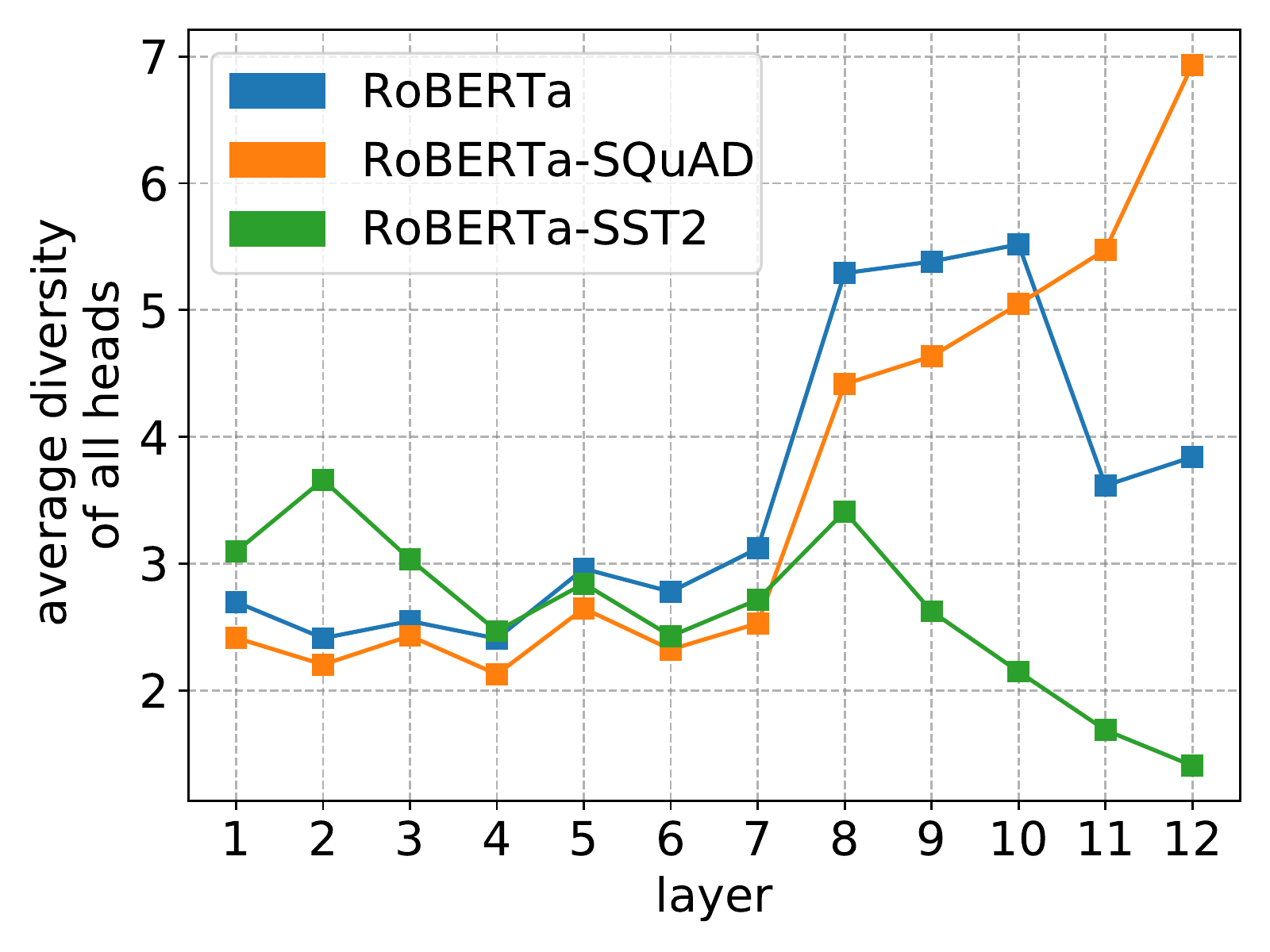} \label{fig:spread_c}}
    \hspace{0.5em}
    \subfloat[][Average dispersion of attention per layer in BERT]
    {\includegraphics[width=0.4\linewidth]{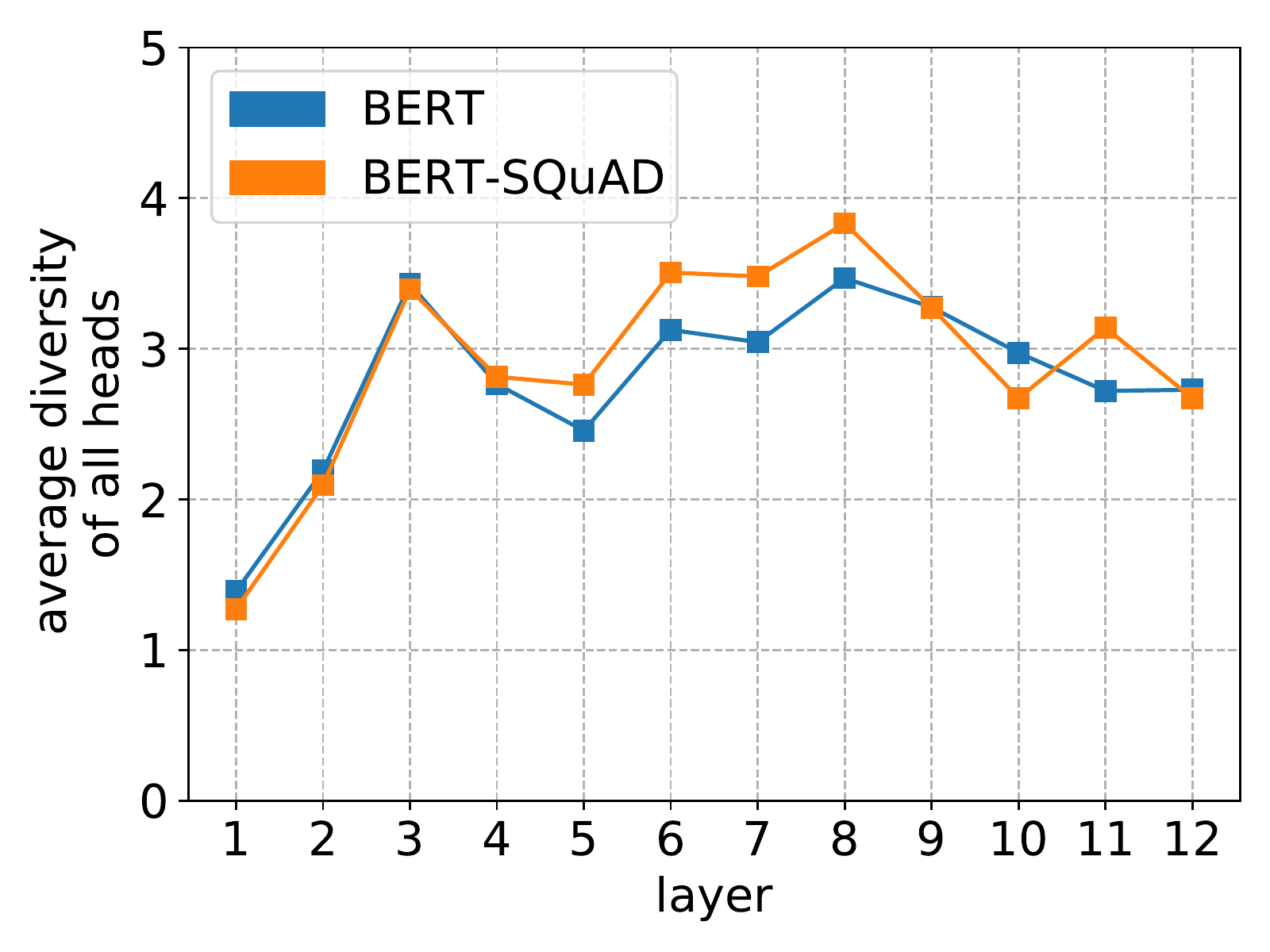} \label{fig:spread_d}}
    \caption{Attention distribution dispersion in different layers. Pretrained RoBERTa has more spread attention distributions in layer 12 than in layer 1. 
    In (c), the pretrained and SQuAD-fine-tuned RoBERTa models exhibit increasing  dispersion in deeper layers, while RoBERTa fine-tuned for SST-2 does not show such a trend.}
    \label{fig:spread}
\end{figure*}

Comparing the attention histograms in the lower layers and the higher layers in RoBERTa (examples shown in Figure~\ref{fig:spread_a} and \ref{fig:spread_b} respectively), we found that the higher layers have more cumulative histograms ``dispersed'' along the x-axis.
Together with the increasing variance of the number of major tokens in the last two layers shown in Figure~\ref{fig:head_consistency},
such a distribution pattern evidently expresses the greatly dissimilar sparsity among all the $\alpha_i$ in the head. 
As a quantitative analysis, we define the \textit{dispersion} of the $\alpha_i$ distribution in a head as the standard deviation of the index of the cumulative histogram bin reaching 0.5.
The dispersion expresses the dissimilarity of the $\alpha_i$ histogram.
Note that this is different from the standard deviation shown in Figure~\ref{fig:head_consistency}, as the dispersion is measuring the histograms of the attention, but not the attention values themselves.

We measure the dispersion at each head along the layers for both pretrained and fine-tuned RoBERTa models. 
Figure~\ref{fig:spread_c} illustrates the changes in dispersion along the layers in the RoBERTa models.
In pretrained RoBERTa and its SQuAD-fine-tuned version, the deep layers generally have higher dispersion.
The difference between these two models is mainly in layer 11, where the pretrained model has a dispersion drop.
RoBERTa fine-tuned for SST-2 does not show this trend.
On the BERT models, dispersion rarely increases along the layers (shown in Figure~\ref{fig:spread_d}).
The last layers have been proved to be task-specific \cite{wu_similarity_2020, rogers_primer_2020}, and their attention can largely change after fine-tuning \cite{kovaleva_revealing_2019}.
This potentially explains why we observed different dispersion behavior on different tasks, but needs further investigation.

\section{Heads with Outlier Attention Distribution}
\label{appx:outlier}

On some heads, a small portion of the tokens forms an attention histogram cluster separate from the majority, clearly showing a dissimilarity between these two types of distributions. 
For example, in Figure~\ref{fig:overview_b}, we observe a small number of tokens clustered on the right of the majority, between $[10^{-4}, 10^{-2}]$.
Here we list all the heads with such pattern:
\begin{itemize}
    \item Pretrained RoBERTa: Layer 1: head 8, head 10, head 12; Layer 2: head 3, head 5, head 10; Layer 3: head 2, head 10; Layer 4: head 4, head 9; Layer 5: head 2, head 7, head 10; Layer 6: head 5, head 11, head 12; Layer 7: head 3; Layer 8: head 7
    \item Pretrained BERT: Layer 3: head 10; Layer 5: head 5
\end{itemize}

\begin{figure}[!ht]
    \centering
    \subfloat[][\centering Layer 2 head 3, \texttt{<s>}, \texttt{le}, \texttt{and}, \texttt{:} and \texttt{</s>} form a weak, less sparse cluster.]
    {\includegraphics[width=0.8\columnwidth]{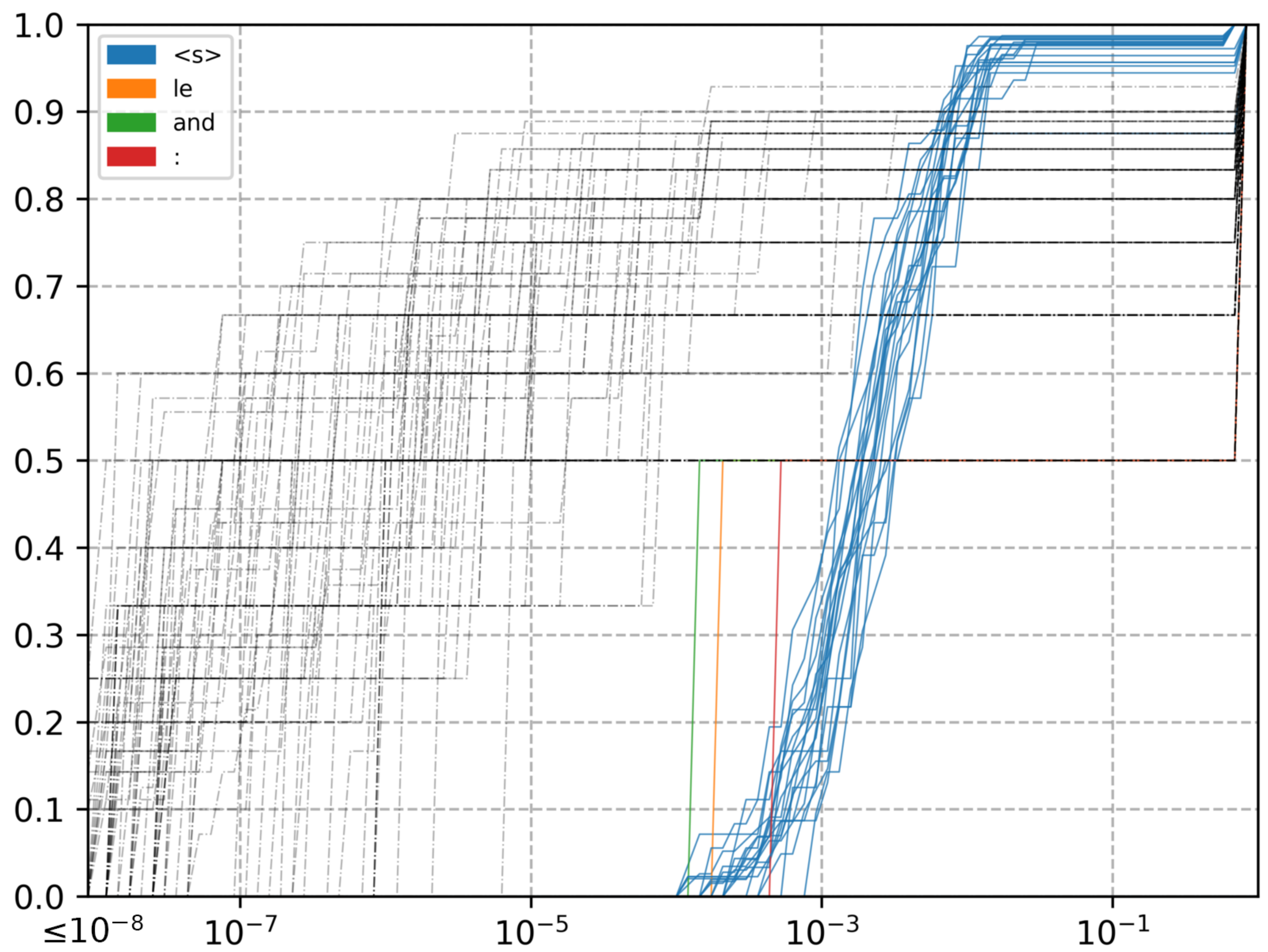} \label{fig:weak_bimodal_a}} \\
    \subfloat[][\centering Layer 4 head 4, \texttt{<s>} and \texttt{.} form a weak, more sparse cluster]
    {\includegraphics[width=0.8\columnwidth]{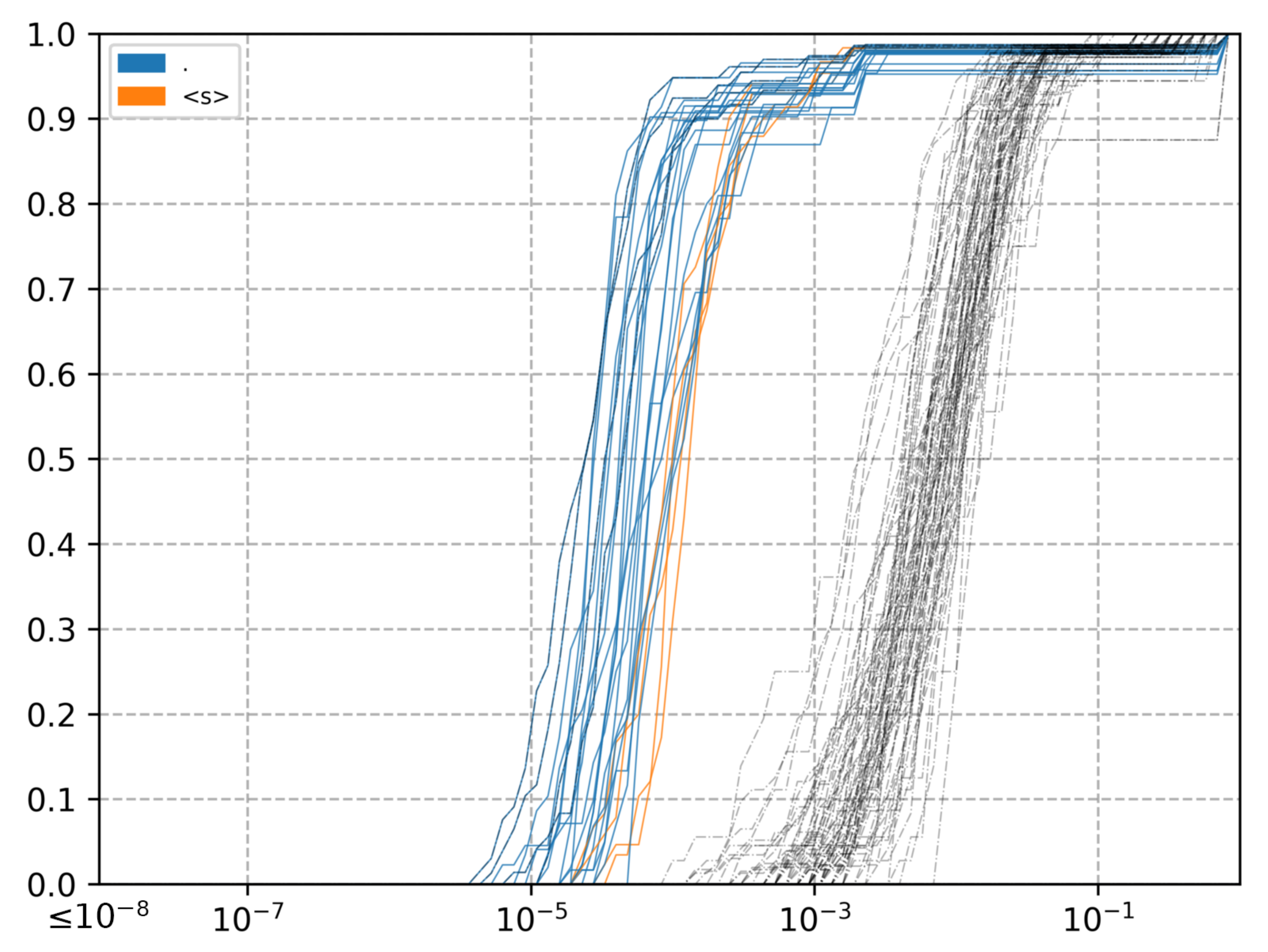} \label{fig:weak_bimodal_b}}
    \caption{A small portion of the tokens cluster outside of the majority of the attention's cumulative  histogram in RoBERTa. Such tokens are noted in different colors with their token strings (\texttt{<s>} and \texttt{</s>} are the ``start of instance'' and ``end of instance'' tokens, respectively), while other tokens are in black as dashed lines. \label{fig:weak_bimodal}}
\end{figure}

We found that on these heads, the functional words/tokens and punctuation exhibit distributions that are significantly different from other tokens. 
For example, tokens such as \texttt{<s>}, \texttt{</s>}, \texttt{and}, \texttt{:} and \texttt{.} are outliers in the pretrained RoBERTa model and \texttt{\lbrack SEP\rbrack} and \texttt{[CLS]} are outliers in the pretrained BERT model.
We also noticed these tokens' attention histograms could gather together like the majority of the tokens do, to form either a less sparse histogram cluster or more sparse histogram cluster, implying that on some heads, the functional words/tokens must be treated differently from the other tokens when exploring efficiency by utilizing sparsity.
In Figure~\ref{fig:weak_bimodal}, we illustrate the attention histogram of such tokens.
Our observation confirms that the special tokens and punctuation can be heavily attended \cite{voita_context-aware_2018, clark_what_2019, kovaleva_revealing_2019, rogers_primer_2020}. 
As a complement, we observed that it does not necessarily mean that the special tokens' attention are always more sparse than other tokens' attention.

\section{Quantization with Pruned Attention for SA and MLM}
\label{appx:quant_sa_mlm}
\begin{figure}[!ht]
    \centering
    \subfloat[][sentiment analysis]
    {\includegraphics[width=0.9\columnwidth]{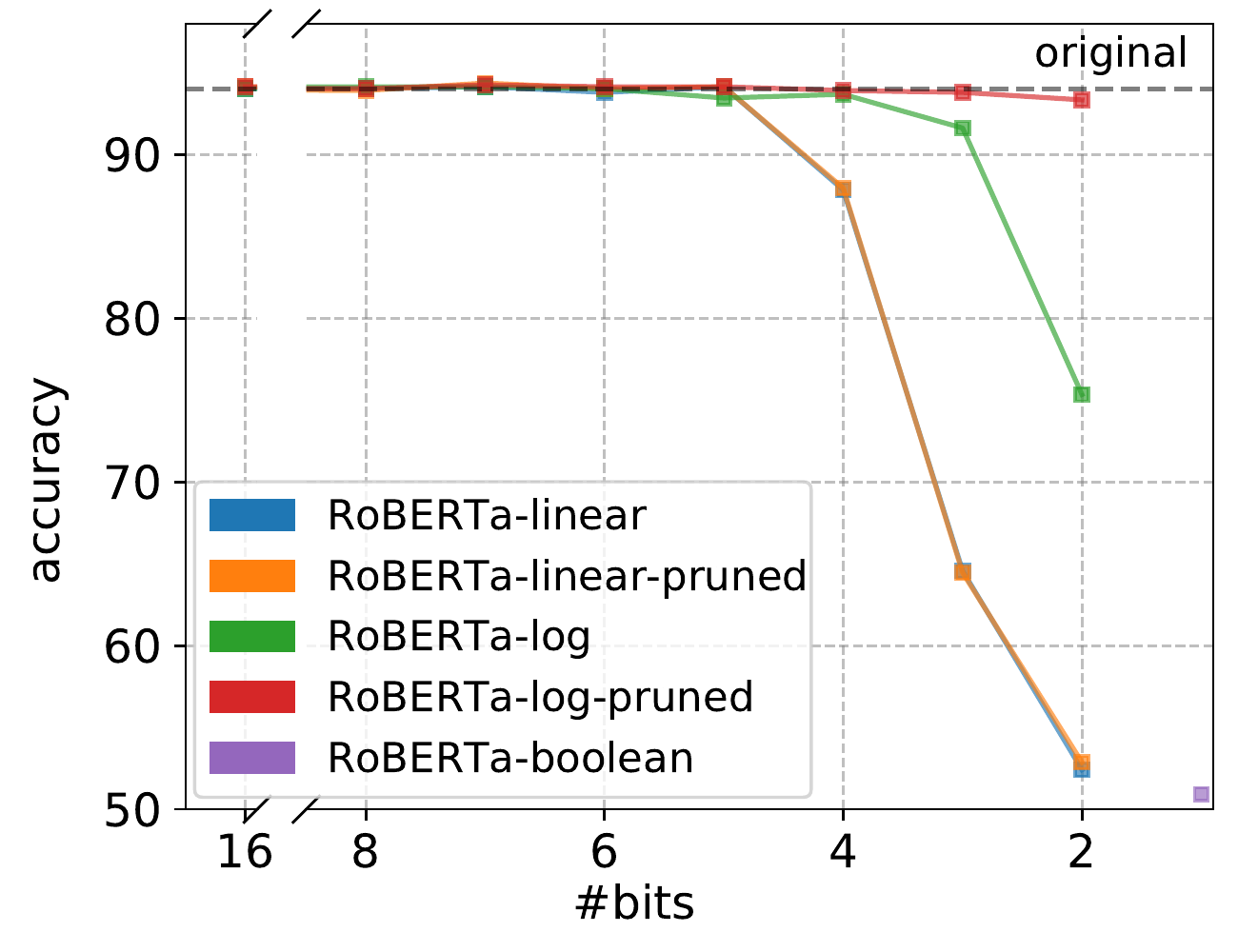} \label{fig:sst_quant}}\\
    \subfloat[][masked language modeling]
    {\includegraphics[width=0.9\columnwidth]{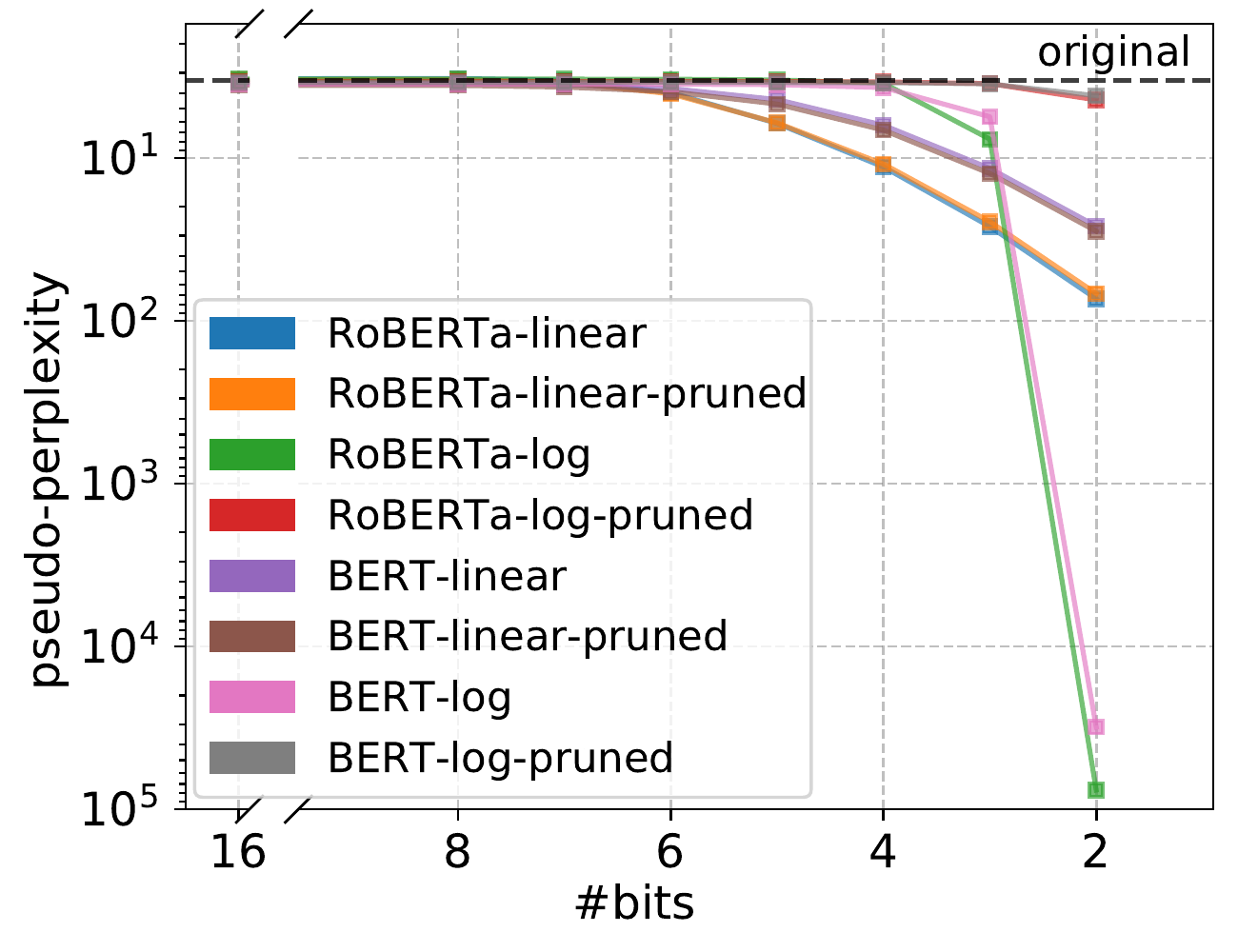} \label{fig:mlm_quant}} 
    \caption{Performance of the quantized models with and without pruning in advance for BERT and RoBERTa models on SA and MLM tasks. \label{fig:additional_quant}}
\end{figure}

We provide the performance of different quantization methods with and without attention pruning on the BERT and RoBERTa models tested on SA and MLM in Figure~\ref{fig:additional_quant}.
\vspace{1em}

\section{Quantization Methods and Their Effectiveness}
\label{appx:quant_method}

\paragraph*{Quantization methods.}
In Section~\ref{subsec:quantize}, we implemented two different quantization methods.
Algorithms~\ref{alg:linear} and \ref{alg:log} list their pseudo code.

\paragraph*{Quantization and attention distribution}

\citet{bhandare_efficient_2019} suggested analyzing the distribution to improve the quantization-effort-intensive functions like softmax (which generates the attention values). 
Based on this, we assume that the transformer model will perform better if its quantized attention values are distributed similarly to the unquantized distribution.
By measuring the average Jensen-Shannon divergence between the original $\alpha_i$ histogram and its quantized version, we found that the logarithmic quantization has lower divergence from the original attention distribution compared to the linear quantization (see Table~\ref{tab:quant_divergence_3bits}).
While in our quantization experiment, the logarithmic quantization indeed achieves higher performance than the linear quantization on most numbers of bits.
This result indicates that selecting the quantization method with less divergence from the original attention distribution could improve the performance.
However, the lower divergence between the quantized and original attention distribution does not necessarily relate to the model performance once we introduce pruning. 
In Table~\ref{tab:quant_divergence_3bits}, even though the histogram's divergence of the pruned log quantization is higher than the un-pruned one, pruning still helps get better results. 
We hypothesize that the pruning enlarged the dissimilarity between the attention histograms, but such a change did not affect the accuracy since it only happened to the near-zero attention values.

\begin{algorithm}[t]
\small
\SetKwInOut{Input}{input}
\SetKwInOut{Output}{output}

\Input{\textit{att}$\leftarrow$attention values;
\newline \textit{k}$\leftarrow$number of bits used for quantization;
\newline \textit{t}$\leftarrow$pruning threshold}
\Output{\textit{res}$\leftarrow$quantized attention values}
quantile\_size = $(1-t)/2^{k}$\;
set quantized\_value as middle point of quantile: quantile\_size/2\;
\textit{res}=floor(\textit{att} / quantile\_size) * quantile\_size + quantized\_value\ + \textit{t}\;
set attention values less than quantile\_size+\textit{t} as zeros; 
\caption{Linear quantization}
\label{alg:linear}
\end{algorithm}

\begin{algorithm}[t]
\small
\SetKwInOut{Input}{input}
\SetKwInOut{Output}{output}

\Input{\textit{att}$\leftarrow$attention values;\newline \textit{k}$\leftarrow$number of bits used for quantization;\newline
\textit{t}$\leftarrow$pruning threshold}
\Output{\textit{res}$\leftarrow$quantized attention values}
when not pruning \textit{att}, choosing a small value $10^{-10}$ for \textit{t}\;
\If {pruning \textit{att}}{
quantile\_size = $(0-\textrm{log}(t))/(2^{k}-1)$\;}
\Else{quantile\_size = $(0-\textrm{log}(t))/(2^{k})$}
set quantized\_value as middle point of quantile: quantile\_size/2\;
compute exponent of \textit{res}: \textit{exp\_res}=floor((log(\textit{att}) $-$ log(\textit{t}))/quantile\_size)*quantile\_size+quantized\_value+\textit{t}\;
\textit{res}=power(2, \textit{exp\_res})\;
set values less than the first quantile boundry in the \textit{res} as zeros; 
\caption{Log quantization}
\label{alg:log}
\end{algorithm}

\begin{table}[t]
\centering
\small
\begin{tabular}{@{}lcc@{}}
\toprule quantization method & pruned & un-pruned \\ \midrule
linear & 0.67 & 0.67 \\
log & 0.58 & 0.55 \\
\bottomrule
\end{tabular}
\caption{Average Jensen-Shannon divergence between the histogram of original $\alpha_i$ and its 3-bit quantized values, evaluated on 100 samples from SQuAD Dev-1.1. Log quantization, which has lower divergence from the original attention distribution, retains more accuracy from the original model.}
\label{tab:quant_divergence_3bits}
\end{table}

\section{Limited Accuracy Change on the Linear Quantization with/without Pruning}
\label{appx:linear_quant_prune}

In Figure~\ref{fig:perf_vs_quant} we observed similar performance of the linear quantized attention models before and after pruning.
It is worth noting that the pruning threshold we selected, $\alpha < 10^{-3}$, is already a tiny value on the linear scale with respect to the range of the attention values $[0, 1]$. 
As a result, pruning will not significantly narrow the quantization range, as it does for the log-scale quantization.
Thus the linear quantization has nearly the same effective quantized range with or without pruning, making it nearly impossible for the pruned linear quantized model to outperform the un-pruned one.
This can be verified by the fact that the Jensen-Shannon Divergence of the linear quantized attention and the original attention's histogram are the same with or without pruning in Table~\ref{tab:quant_divergence_3bits}.

\section{Experiment reproducibility}
All evaluation is done on a server with the following specifications:
\begin{itemize}
    \item CPU: Intel(R) Xeon(R) Silver 4216, 64 cores
    \item GPU: Quadro RTX 8000
    \item RAM: 377GB
\end{itemize}
The average runtime of the model inferences through the entire dataset is $\sim$4 hours, for different tasks.
All datasets used in our experiment are based on English.
The SQuAD tests are evaluated on 10570 sentences from the SQuAD Dev-v1.1 dataset.
The SST2 tests are evaluated on 872 instances from the GLUE validation dataset.
The Masked Language Modeling tests are evaluated on 480 paragraphs from the wikipedia training set, each having one random, unrepeated token masked for 15--25 iterations.

\end{document}